\documentclass{article}
\usepackage{spconf,amsmath,epsfig,amssymb}
\usepackage{multirow}
\usepackage{booktabs}
\usepackage{graphicx, color}
\usepackage[normalem]{ulem}
\usepackage[switch, mathlines]{lineno}
\usepackage{comment}
\usepackage{hyperref}

\let\OLDthebibliography\thebibliography
\renewcommand\thebibliography[1]{
  \OLDthebibliography{#1}
  \setlength{\parskip}{0pt}
  \setlength{\itemsep}{0pt plus 0.3ex}
}

\begin{document}\sloppy
%\linenumbers 

% Title.
% ------
\title{Can SAM Count Anything? An Empirical Study on SAM Counting}
%\title{XXXX: An Efficient Network For One-Shot Object Counting}
%
% Single address.
% ---------------
\name{Zhiheng Ma$^{1}$, Xiaopeng Hong$^{2,3}$, Qinnan Shangguan$^{2}$}
\address{$^{1}$Shenzhen Institute of Advanced Technology, Chinese Academy of Science  \\ $^{2}$Harbin Institute of Technology, Harbin, P. R. China \\ $^3$Peng Cheng Laboratory, Shenzhen, P. R. China}
%\name{}
%\small Emails: {Emails: ;  hongxiaopeng@ieee.org; }
%Address and e-mail should NOT be added in the submission paper. They should be present only in the camera ready paper. 
%\address{\small School of Cyber Science and Engineering, Xi'an Jiaotong University, China\\

\maketitle

\begin{abstract}

Meta AI recently released the Segment Anything model (SAM), which has garnered attention due to its impressive performance in class-agnostic segmenting. In this study, we explore the use of SAM for the challenging task of few-shot object counting, which involves counting objects of an unseen category by providing a few bounding boxes of examples. We compare SAM's performance with other few-shot counting methods and find that it is currently unsatisfactory without further fine-tuning, particularly for small and crowded objects. Code can be found at \url{https://github.com/Vision-Intelligence-and-Robots-Group/count-anything}.

\end{abstract}
%

% \begin{figure*}[t]
%     \centering
%     \includegraphics[width = 0.95\textwidth]{fig/network.png}
%     \vspace{-4mm}
%     \caption{The overall architecture of the proposed LaoNet for one-shot object counting. Both the query image and the supporting box are fed into CNN to extract features. Supporting features are aggregated among scales. Then the flatten features with unique position embedding are transmitted into feature correlation model with Self-Attentions and Correlative Attentions. Finally, a density regressor is adopt\XP{ed} to predict the final density map. }
%     \label{fig:network}
%     \vspace{-4mm}
% \end{figure*}

% -------------------------------------------------------------------------

\section{Introduction}

Large models have brought about a revolution in the field of AI, transforming it in numerous ways. In the past few months, Natural Language Processing (NLP) has undergone a significant shift towards the development of large-scale language models, resulting in some remarkable applications such as ChatGPT~\cite{brown2020language} and GPT-4~\cite{openai2023gpt4}. The success of large models in NLP has also inspired researchers to explore their application in the field of Computer Vision (CV).

Recently, Meta AI has released the Segment Anything model (SAM)~\cite{kirillov2023segment}, which was trained on over 1 billion masks using 11 million licensed and privacy-respecting images. Its exceptional performance in segmenting unknown images has rapidly gained attention. SAM has demonstrated its impressive capabilities in segmenting various types of images and scenes. We are therefore excited to investigate SAM's performance in few-shot counting to further explore its potential.

In this report, we compare SAM with other existing few-shot counting methods. Whether SAM can successfully count target objects depends on two aspects: first, whether SAM can segment each  object. Second, whether SAM can distinguish the target objects from others using the reference examples. Instead of introducing an additional zero-shot object detector such as Grounding DINO~\cite{ShilongLiu2023GroundingDM}, or an additional zero-shot classifier such as CLIP~\cite{radford2021learning}, we use the original image features of SAM to distinguish different objects, which greatly save the computational cost.

We provide a detailed account of our implementation as follows. Firstly, we extract the dense image feature through the use of the image encoder (ViT-H) of SAM for a given image. Secondly, we utilize the given bounding boxes as prompts to generate segment masks of the reference examples. These masks are then multiplied with the dense image feature and subsequently averaged to produce the feature vectors of the reference objects. Thirdly, we use the point grids (32 points on each side) as prompts to segment everything, and the output masks are multiplied with the dense image feature before being averaged to generate feature vectors of all masks. Finally, we compute the cosine similarity between the feature vectors of the predicted masks and the reference examples. If the cosine similarity exceeds the predetermined threshold, we consider it as the targeted object. The total count can then be obtained by counting all the target objects.

\section{Experiments}

\subsection{Datesets}

\noindent\textbf{FSC-147}~\cite{ranjan2021learning} comprises a total of 6,135 images that have been collected for the purpose of few-shot counting. For each of these images, three object instances are randomly selected and annotated using bounding boxes, while the remaining instances are annotated using points. The training set consists of 89 object categories with 3,659 images. Meanwhile, the validation and testing sets each consist of 29 categories, with 1,286 and 1,190 images respectively.

\noindent\textbf{MS-COCO}~\cite{lin2014microsoft} is a large dataset that is commonly used in object detection and instance segmentation. The val2017 set contains 5,000 images of 80 object categories that are present in complex everyday scenes. We adopt the same approach as in~\cite{michaelis2018one} and generate four train/test splits, each of which contains 60 training categories and 20 testing categories. It is noteworthy that while other methods utilize the training set for fine-tuning, SAM refrains from doing so.

\renewcommand{\tabcolsep}{10 pt}{
\begin{table*}[!t]
\small
\begin{center}
\begin{tabular}{lcccccccccc}
  \toprule[1pt]
  \multirow{2}*{Methods} & \multicolumn{2}{c}{Fold\ 0} & \multicolumn{2}{c}{Fold\ 1} & \multicolumn{2}{c}{Fold\ 2} & \multicolumn{2}{c}{Fold\ 3} & \multicolumn{2}{c}{Average}\\
  & MAE & RMSE & MAE & RMSE & MAE & RMSE & MAE & RMSE & MAE & RMSE\\
  \hline
  Segment~\cite{michaelis2018one}$^\dag$ & 2.91 & 4.20 & 2.47 & 3.67 & 2.64 & 3.79 & 2.82 & 4.09 & 2.71 & 3.94 \\
  GMN~\cite{lu2018class}$^\dag$ & 2.97 & 4.02 & 3.39 & 4.56 & 3.00 & 3.94 & 3.30 & 4.40 & 3.17 & 4.23 \\
  CFOCNet~\cite{yang2021class}$^\dag$ & 2.24 & \textbf{3.50} & 1.78 & 2.90 & 2.66 & 3.82 & 2.16 & 3.27 & 2.21 & 3.37 \\
  \hline
  FamNet~\cite{ranjan2021learning} & 2.34 & 3.78 & 1.41 & 2.85 & 2.40 & 2.75 & 2.27 & 3.66 & 2.11 & 3.26 \\
  CFOCNet~\cite{yang2021class} & 2.23 & 4.04 & 1.62 & 2.72 & 1.83 & 3.02 & 2.13 & 3.03 & 1.95 & 3.20 \\
  LaoNet~\cite{lin2021object} & \textbf{2.20} & 3.78 & \textbf{1.32} & \textbf{2.66} & \textbf{1.58} & \textbf{2.19} & \textbf{1.84} & \textbf{2.90} & \textbf{1.73} & \textbf{2.93} \\
  SAM & 5.27 & 7.67 & 3.85 & 8.51 & 3.13 & 8.09 & 3.23 & 7.87 & 3.87 & 8.03 \\
  \toprule[1pt] \\
\end{tabular}
\caption{The results for each of the four folds of COCO val2017 are presented above.Methods marked with $\dag$ follow the experimental settings in~\cite{yang2021class}.} \label{tab:coco}
\end{center}
\end{table*}}

\renewcommand{\tabcolsep}{8 pt}{
\begin{table}[t]
\small
\begin{center}
\begin{tabular}{lcccc}
  \toprule[1pt]
  \multirow{2}*{Methods} & \multicolumn{2}{c}{Val} & \multicolumn{2}{c}{Test}\\
  & MAE & RMSE & MAE & RMSE \\
  \hline
  \textit{3-shot} & & & & \\
  Mean & 53.38 & 124.53 & 47.55 & 147.67 \\
  Median & 48.68 & 129.70 & 47.73 & 152.46 \\
  FR detector~\cite{kang2019few} & 45.45 & 112.53 & 41.64 & 141.04 \\
  FSOD detector~\cite{fan2020few} & 36.36 & 115.00 & 32.53 & 140.65 \\
%   Pre-trained GMN & 60.56 & 137.78 & 62.69 & 159.67 \\
  GMN~\cite{lu2018class} & 29.66 & 89.81 & 26.52 & 124.57 \\
  MAML~\cite{finn2017model} & 25.54 & 79.44 & 24.90 & 112.68 \\
  FamNet~\cite{ranjan2021learning} & 23.75 & 69.07 & 22.08 & 99.54 \\
  CFOCNet~\cite{yang2021class} & 21.19 & 61.41 & 22.10 & 112.71 \\
  BMNet+~\cite{Shi_2022_CVPR} & 15.74 & 58.53 & 14.62 & 91.83 \\
  SAFECount~\cite{You_2023_WACV} & 15.28 & 47.20 & 14.32 & 85.54 \\
  HMFENet~\cite{10.3389/fncom.2023.1145219} & \textbf{13.10} & \textbf{44.90} & \textbf{12.74} & \textbf{84.63} \\
  SAM & 31.20 & 100.83 & 27.97 & 131.24 \\
  \hline
 \textit{1-shot} & & & & \\
  CFOCNet~\cite{yang2021class} & 27.82 & 71.99 & 28.60 & 123.96 \\
  FamNet~\cite{ranjan2021learning} & 26.55 & 77.01 & 26.76 & 110.95 \\
  LaoNet~\cite{lin2021object} & 17.11 & 56.81 & 15.78 & 97.15 \\
  SAM & 36.68 & 116.75 & 33.53 & 142.28 \\
  \toprule[1pt]
\end{tabular}
\caption{The table displays a comparison between our few-shot method and previous state-of-the-art methods on FSC-147  dataset. The top part of the table shows the results for the 3-shot setting, while the bottom part shows the results for the 1-shot setting. } \label{tab:fam}
\end{center}
\end{table}}

\renewcommand{\tabcolsep}{4 pt}{
\begin{figure*}[t!]
	\begin{center}
		\begin{tabular}{cccc}
			\includegraphics[height=0.22\linewidth]{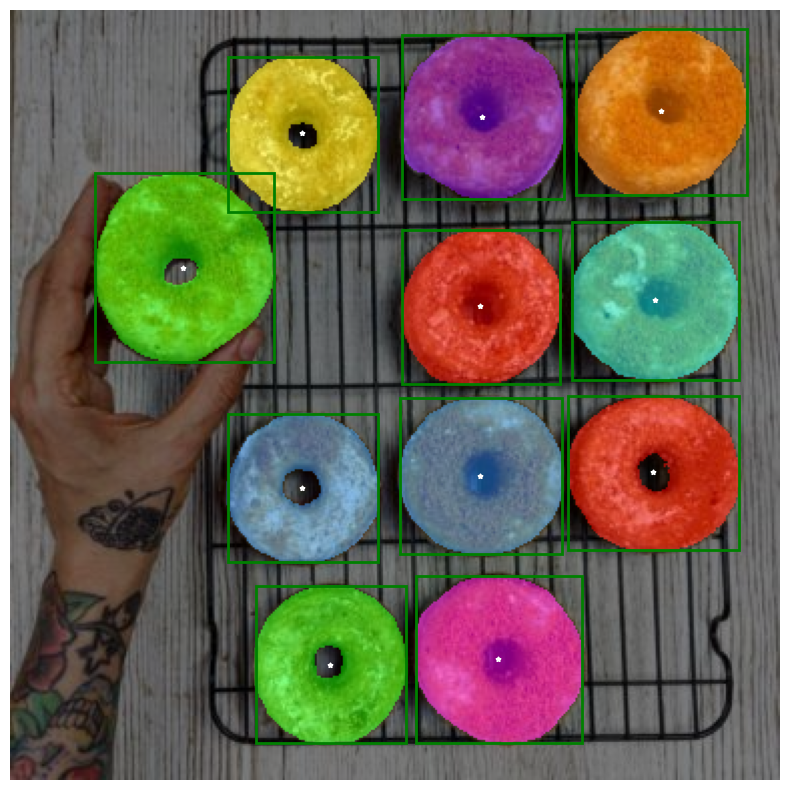}  &
			\includegraphics[height=0.22\linewidth]{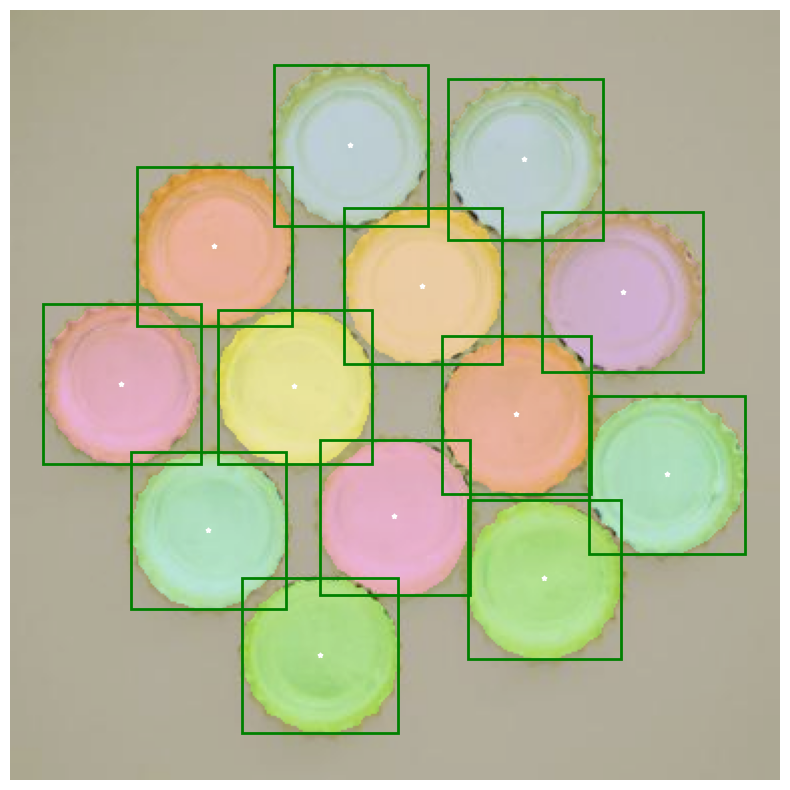}  &
			\includegraphics[height=0.22\linewidth]{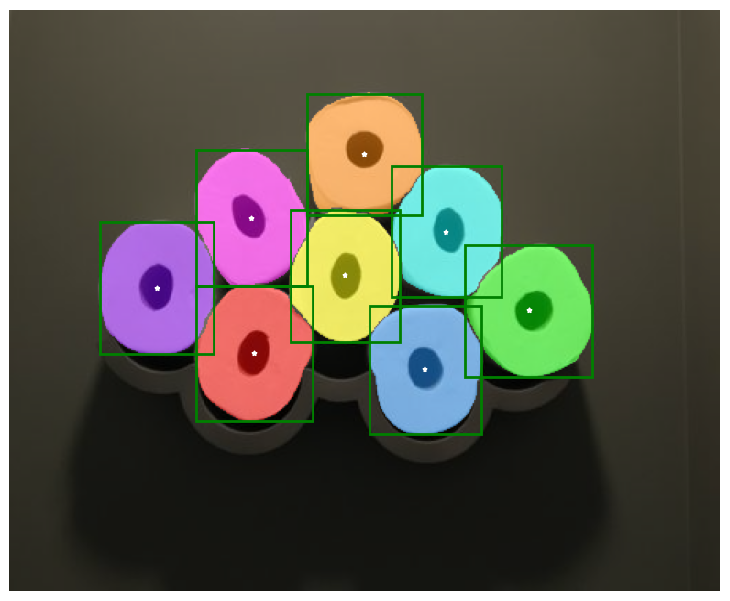}  &
                \includegraphics[height=0.22\linewidth]{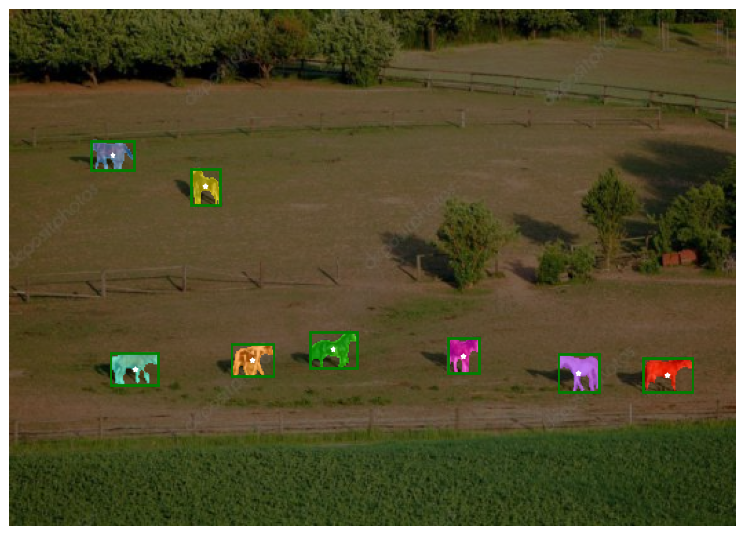}   \\
			
			\includegraphics[height=0.22\linewidth]{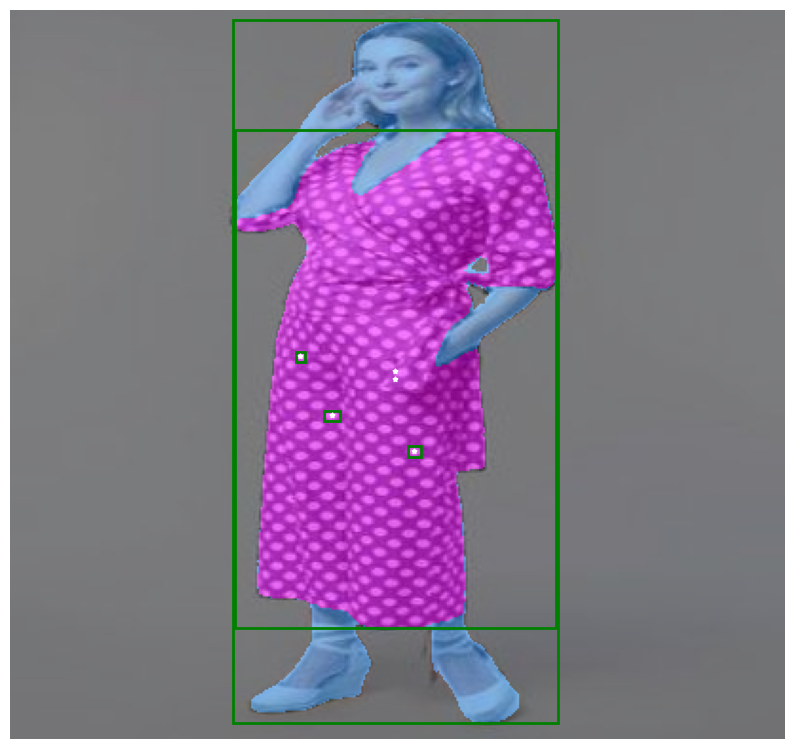}  &
			\includegraphics[height=0.22\linewidth]{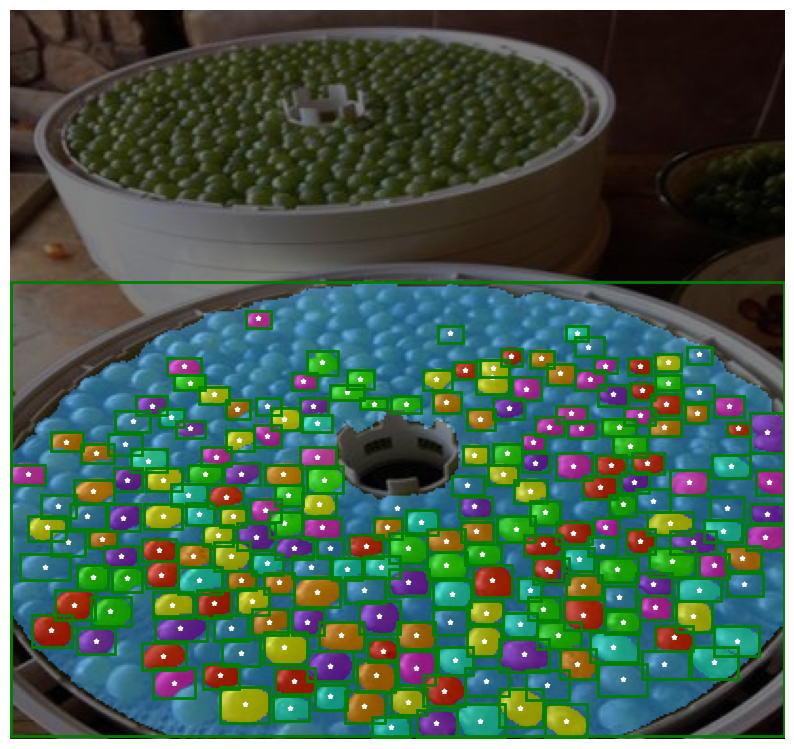}  &
			\includegraphics[height=0.22\linewidth]{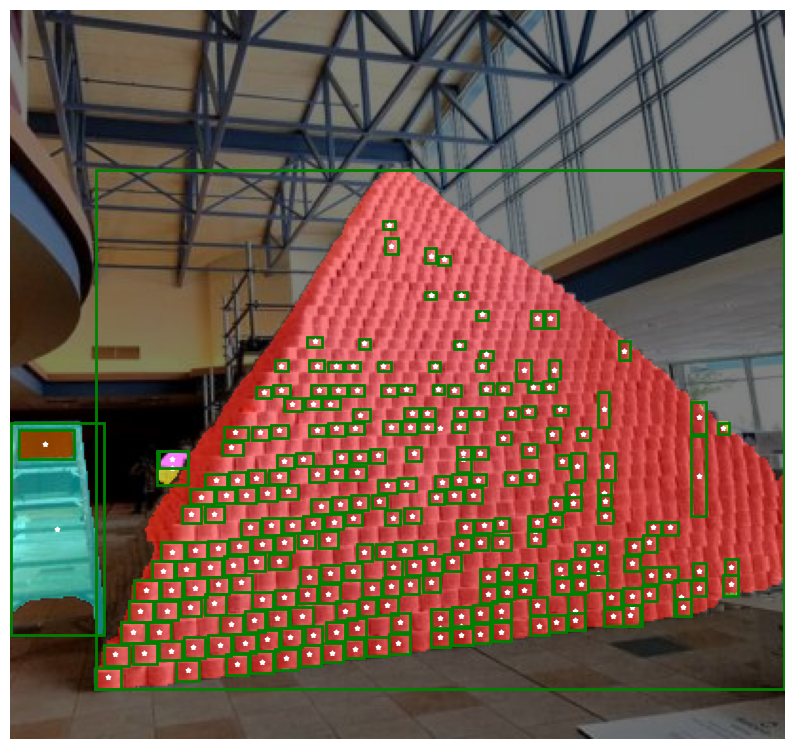}  &
                \includegraphics[height=0.22\linewidth]{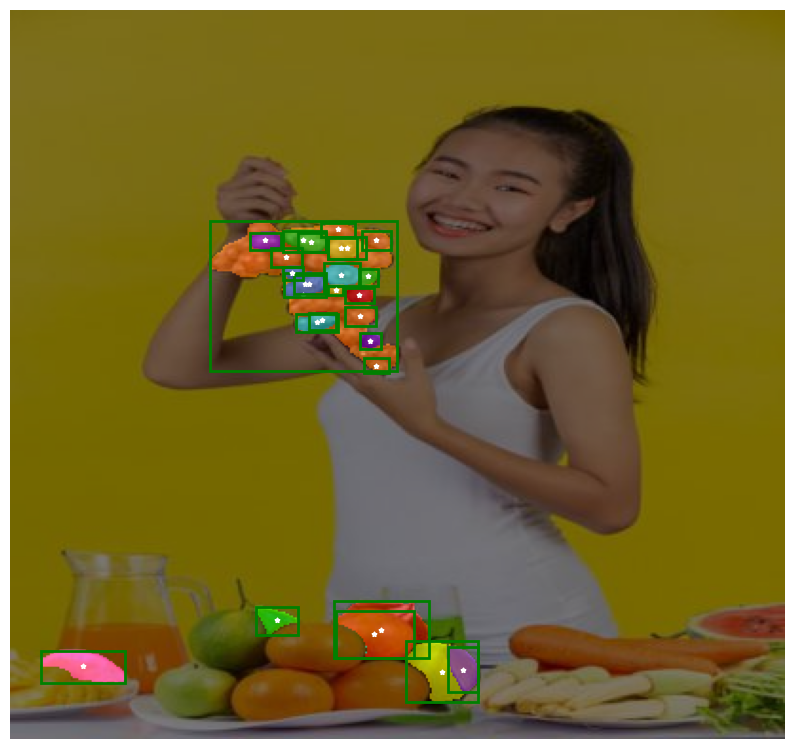}    \\

		\end{tabular}
		\caption{Visualizations of the resulting mask obtained through SAM segmentation and counting are presented. The counts of the four images in the first row are exactly accurate, while the results of the four images in the second row are suboptimal. All the examples are from the FSC-147 dataset.}
		\label{fig:vis}
	\end{center}
\end{figure*}}

\subsection{Comparison with Few-Shot Approaches} \label{sec:comparison}

SAM's performance was evaluated through experiments conducted on the two few-shot counting datasets mentioned earlier. Table~\ref{tab:coco} presents the results of a four-fold cross-validation performed on the COCO val2017 dataset. Methods marked with $\dag$ in the upper part of the table adhere to the experimental setup outlined in ~\cite{yang2021class}. While SAM's performance is comparatively inferior to that of other few-shot methods, we observe that the performance gap is \emph{not significant}, with an about 2-unit gap in average MAE, owing to the fact that COCO contains fewer small and congested objects.

Table~\ref{tab:fam} presents the performance of SAM on the FSC-147 datasets. While SAM's performance surpasses that of some earlier methods, it is \emph{significantly inferior} to that of more recent methods, with a gap of over 10 in MAE. We conducted a visualization study on the FSC dataset, and the results are presented in Figure~\ref{fig:vis}. The first row of images in the figure depicts count predictions that are exactly accurate, and we observe that the items to be counted in these images are relatively sparse. Conversely, the second row of images represents some of the bad examples, where the predicted count and ground truth count differ significantly. The first three images in this row have one thing in common - small and crowded objects are predicted as a single object during prediction. In the fourth image, SAM erroneously predicts other categories of fruit as grapes.

% We hold experiments on above two few-shot counting datasets to evaluate the proposed network. First, Table~\ref{tab:coco} shows the results on each of four folds of COCO val2017. Methods with $\dag$ in the upper part of the table follow the experiment setting in ~\cite{yang2021class}. That is, the supporting examples are chosen from all instances in the dataset during training and testing, which is laborious and costly under the need of all instances annotated by bounding boxes. While our setting allows only one fixed instance for each image, we re-conduct the experiment of CFOCNet~\cite{yang2021class}. As the result shows, our method maintains a great performance on COCO dataset.

%As this study is the first one for one-shot counting and no existing study is provided exactly in this field, 

% Table~\ref{tab:fam} presents the quantitative results on FSC-147. We provide a comparison with state-of-the-art results of previous few-shot detection and counting methods in both 3-shot and 1-shot settings. It is worth noting that the result of FamNet~\cite{ranjan2021learning} utilizes an adaptation strategy during testing. 

% -------------------------------------------------------------------------
\section{Conclusion}

Although the Segment Anything model (SAM) has shown impressive performance in many scenarios, it currently lags behind state-of-the-art few-shot counting methods, especially for small and congested objects. We believe that this is due to two main reasons. Firstly, SAM tends to segment congested objects of the same category with a single mask. Secondly, SAM is trained with masks that lack semantic class annotations, which could hinder its ability to differentiate between different objects. Nevertheless, further exploration of adapting SAM to the object counting task is still worth studying.

\section*{Acknowledgements}
This work is funded by the National Natural Science Foundation of China (62206271, 62076195), the Guangdong Basic and Applied Basic Research Foundation (2020B1515130004), and the Fundamental Research Funds for the Central Universities (AUGA5710011522).
% -------------------------------------------------------------------------
\bibliographystyle{IEEEbib}
\bibliography{ref}

\begin{thebibliography}{10}

\bibitem{brown2020language}
Tom Brown, Benjamin Mann, Nick Ryder, Melanie Subbiah, Jared~D Kaplan, Prafulla
  Dhariwal, Arvind Neelakantan, Pranav Shyam, Girish Sastry, Amanda Askell,
  et~al.,
\newblock ``Language models are few-shot learners,''
\newblock {\em Advances in neural information processing systems}, vol. 33, pp.
  1877--1901, 2020.

\bibitem{openai2023gpt4}
OpenAI,
\newblock ``Gpt-4 technical report,'' 2023.

\bibitem{kirillov2023segment}
Alexander Kirillov, Eric Mintun, Nikhila Ravi, Hanzi Mao, Chloe Rolland, Laura
  Gustafson, Tete Xiao, Spencer Whitehead, Alexander~C. Berg, Wan-Yen Lo, Piotr
  Dollár, and Ross Girshick,
\newblock ``Segment anything,'' 2023.

\bibitem{ShilongLiu2023GroundingDM}
Shilong Liu, Zhaoyang Zeng, Tianhe Ren, Feng Li, Hao Zhang, Jie Yang, Chunyuan
  Li, Jianwei Yang, Hang Su, Jun Zhu, and Lei Zhang,
\newblock ``Grounding dino: Marrying dino with grounded pre-training for
  open-set object detection,''
\newblock 2023.

\bibitem{radford2021learning}
Alec Radford, Jong~Wook Kim, Chris Hallacy, Aditya Ramesh, Gabriel Goh,
  Sandhini Agarwal, Girish Sastry, Amanda Askell, Pamela Mishkin, Jack Clark,
  Gretchen Krueger, and Ilya Sutskever,
\newblock ``Learning transferable visual models from natural language
  supervision,'' 2021.

\bibitem{ranjan2021learning}
Viresh Ranjan, Udbhav Sharma, Thu Nguyen, and Minh Hoai,
\newblock ``Learning to count everything,''
\newblock in {\em CVPR}, 2021.

\bibitem{lin2014microsoft}
Tsung-Yi Lin, Michael Maire, Serge Belongie, James Hays, Pietro Perona, Deva
  Ramanan, Piotr Doll{\'a}r, and C~Lawrence Zitnick,
\newblock ``Microsoft coco: Common objects in context,''
\newblock in {\em ECCV}. Springer, 2014.

\bibitem{michaelis2018one}
Claudio Michaelis, Ivan Ustyuzhaninov, Matthias Bethge, and Alexander~S Ecker,
\newblock ``One-shot instance segmentation,''
\newblock {\em arXiv preprint}, 2018.

\bibitem{lu2018class}
Erika Lu, Weidi Xie, and Andrew Zisserman,
\newblock ``Class-agnostic counting,''
\newblock in {\em ACCV}, 2018.

\bibitem{yang2021class}
Shuo-Diao Yang, Hung-Ting Su, Winston~H Hsu, and Wen-Chin Chen,
\newblock ``Class-agnostic few-shot object counting,''
\newblock in {\em WACV}, 2021.

\bibitem{lin2021object}
Hui Lin, Xiaopeng Hong, and Yabin Wang,
\newblock ``Object counting: You only need to look at one,'' 2021.

\bibitem{kang2019few}
Bingyi Kang, Zhuang Liu, Xin Wang, Fisher Yu, Jiashi Feng, and Trevor Darrell,
\newblock ``Few-shot object detection via feature reweighting,''
\newblock in {\em ICCV}, 2019.

\bibitem{fan2020few}
Qi~Fan, Wei Zhuo, Chi-Keung Tang, and Yu-Wing Tai,
\newblock ``Few-shot object detection with attention-rpn and multi-relation
  detector,''
\newblock in {\em CVPR}, 2020.

\bibitem{finn2017model}
Chelsea Finn, Pieter Abbeel, and Sergey Levine,
\newblock ``Model-agnostic meta-learning for fast adaptation of deep
  networks,''
\newblock in {\em ICML}, 2017.

\bibitem{Shi_2022_CVPR}
Min Shi, Hao Lu, Chen Feng, Chengxin Liu, and Zhiguo Cao,
\newblock ``Represent, compare, and learn: A similarity-aware framework for
  class-agnostic counting,''
\newblock in {\em Proceedings of the IEEE/CVF Conference on Computer Vision and
  Pattern Recognition (CVPR)}, June 2022, pp. 9529--9538.

\bibitem{You_2023_WACV}
Zhiyuan You, Kai Yang, Wenhan Luo, Xin Lu, Lei Cui, and Xinyi Le,
\newblock ``Few-shot object counting with similarity-aware feature
  enhancement,''
\newblock in {\em Proceedings of the IEEE/CVF Winter Conference on Applications
  of Computer Vision (WACV)}, January 2023, pp. 6315--6324.

\bibitem{10.3389/fncom.2023.1145219}
Zhiquan He, Donghong Zheng, and Hengyou Wang,
\newblock ``Accurate few-shot object counting with hough matching feature
  enhancement,''
\newblock {\em Frontiers in Computational Neuroscience}, vol. 17, 2023.

\end{thebibliography}

\end{document}